\pdfoutput=1
\documentclass[11pt]{article}

\usepackage{naacl2021}

\usepackage{times}
\usepackage{latexsym}

\usepackage[T1]{fontenc}
\usepackage[utf8]{inputenc}

\usepackage{microtype}

\title{Multi-Task Learning of Generation and Classification \\
       for Emotion-Aware Dialogue Response Generation}

\author{Tatsuya Ide \and Daisuke Kawahara \\
        Waseda University \\
        \texttt{\{t-ide@toki., dkw@\}waseda.jp}}

\usepackage{bm}
\usepackage{graphicx}
\usepackage{url}
\usepackage{multirow}

\begin{document}
\maketitle
\begin{abstract}
For a computer to naturally interact with a human, it needs to be human-like.
In this paper, we propose a neural response generation model with multi-task learning of generation and classification, focusing on emotion.
Our model based on BART~\citep{lewis-etal-2020-bart}, a pre-trained transformer encoder-decoder model, is trained to generate responses and recognize emotions simultaneously.
Furthermore, we weight the losses for the tasks to control the update of parameters.
Automatic evaluations and crowdsourced manual evaluations show that the proposed model makes generated responses more emotionally aware.
\end{abstract}

\section{Introduction}

The performance of machine translation and summarization has been approaching a near-human level in virtue of pre-trained encoder-decoder models, such as BART~\citep{lewis-etal-2020-bart} and T5~\citep{JMLR:v21:20-074}.
The same technology has been applied to dialogue systems, which are now expected to be put to practical use.

To interact naturally with a human, the computer needs to be human-like.
Several methods have been proposed to build such dialogue systems.
They include a system interacting based on knowledge and common sense~\citep{dinan2019wizard} and that interacting by considering one's own and the other's personality~\citep{zhang-etal-2018-personalizing}.
In particular, we focus on the viewpoint of emotion as targeted in \citet{rashkin-etal-2019-towards}.

In this paper, we propose a multi-task learning method for building a dialogue system that takes the speaker's emotions into account.
Also, we focus on the hierarchy of emotions~\citep{8852352} and simultaneously train multiple emotion recognition tasks with different granularity.
Our multi-task learning model is not expected to share complementary information among similar tasks as previous work~\citep{liu-etal-2019-multi-task}, and we do not aim at improving the accuracy of emotion recognition. Instead, we focus on generating emotion-aware responses.
Also, concerned that the ratio of emotion recognition in multi-task learning is too large, we explore further quality improvement by weighting each loss.
We build a model based on BART~\citep{lewis-etal-2020-bart}, a pre-trained Transformer~\citep{vaswani2017attention} model, to implement multi-task learning of response generation and emotion recognition.

Experiments are performed using a dialogue corpus without context.
The effectiveness of the proposed method in generating responses is confirmed by automatic and manual evaluations.
Multi-task learning of response generation and emotion recognition makes generated responses more emotionally aware of utterances.
The improvement is not only on the emotional aspect but also on the quality of fluency, informativeness, and relevance.
We also found that controlling the parameters by weighting the losses improved the performance of the model.

\begin{figure*}[t]
    \centering
    \includegraphics[width=.8\linewidth]{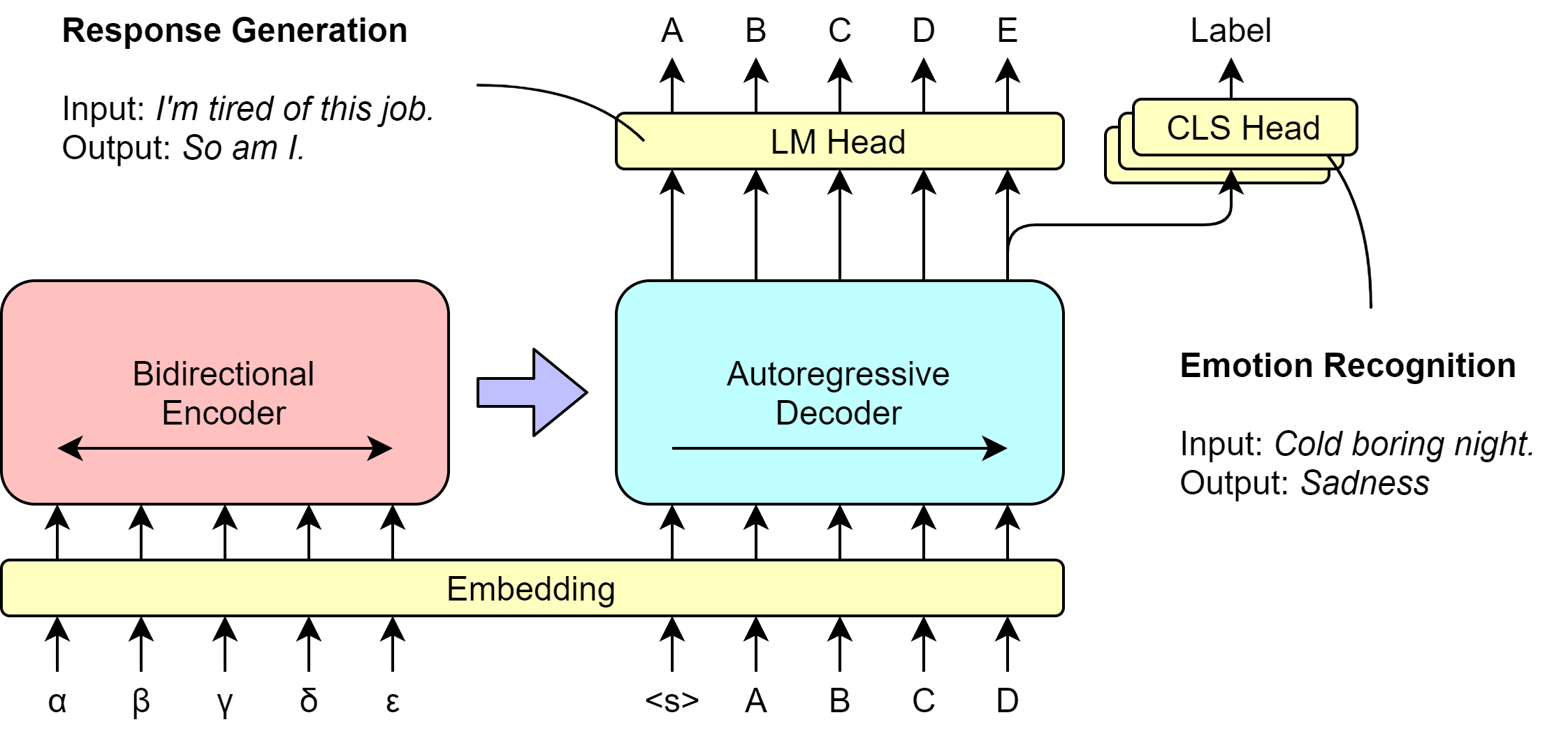}
    \caption{The architecture of our model, based on BART~\citep{lewis-etal-2020-bart}.
             It contains one LM head and several CLS heads, which solve generation and classification, respectively. In our experiments, three CLS heads are used for the emotion recognition tasks with different granularity.}
    \label{fig:arch}
\end{figure*}

\section{Related Work}

One of the previous studies on emotion-based response generation is the Emotional Chatting Machine (ECM)~\citep{Zhou_Huang_Zhang_Zhu_Liu_2018}.
ECM is used together with an emotion classifier to generate a response based on a given emotion.
EmpTransfo~\citep{zandie2020emptransfo} is a similar model to ours.
Given an utterance, a model based on GPT~\citep{radford2018improving} learns an emotion and an action simultaneously in addition to a response, which improves the quality of generated responses.
These models focus on the emotion of a response so that they do not generate a response based on the emotion of an utterance.

\citet{Lubis_Sakti_Yoshino_Nakamura_2018} incorporate an emotion encoder into a hierarchical seq2seq architecture, enabling a system to understand the emotional context on a user.
TG-EACM~\citep{wei2020target}, the successor of EACM~\citep{10.1145/3357384.3357937}, is a model that considers not only the emotion in an utterance but also the emotion that a response should have.
The model learns a distribution to infer both the emotion of the utterance and the response from a given utterance.
CARE~\citep{zhong2021care} uses some commonsense to generate a response with both rationality and emotion.
Through latent concepts obtained from an emotionally aware knowledge graph, predicted responses can be emotional and rational.

Actually, the above models require separate units or special architecture for understanding emotion in a dialogue.
In contrast, our proposed model achieves that with a single structure, inherited from Transformer~\citep{vaswani2017attention} and BART~\citep{lewis-etal-2020-bart}.
In other words, our model does not need an extra unit.
Therefore, the proposed method consequently reduces the redundancy of Transformer parameters~\citep{kovaleva-etal-2019-revealing} and realizes more efficient understanding of emotion to generate a response.

\section{Emotion-Aware Response Generation by Multi-Task Learning}

\subsection{Overview}

Our model learns response generation as a generation task and emotion recognition as a classification task.
By learning response generation and emotion recognition simultaneously through multi-task learning, it is possible to generate a response by considering the emotion of a given utterance.

Multi-task learning often involves several similar tasks because they can share information and thus the performance of each task can be improved.
However, the purpose of our multi-task learning method is to improve the quality of response generation, not to improve the performance of emotion recognition.
This is different from general multi-task learning.

Our model is based on BART~\citep{lewis-etal-2020-bart}. Its architecture is shown in Figure~\ref{fig:arch}.
The model has several output layers, or heads, for the tasks to be trained, which include an LM head for generating words in response generation and CLS heads for solving classification tasks.
Given a sentence, the CLS head predicts its label such as \texttt{positive} or \texttt{negative}.
One CLS head is set for each classification task.

The input/output format of each task is the same as that in BART.
In the generation task, we put an utterance and a right-shifted response into the encoder and decoder, respectively.
In the classification task, we put an utterance and a right-shifted utterance into the encoder and decoder, respectively.
Following the learning algorithm of MT-DNN~\citep{liu-etal-2019-multi-task}, each task that the model learns is selected for each mini-batch.
A different loss is calculated for each task, and the parameters are updated for each mini-batch.

\subsection{Losses of Generation and Classification Tasks}

Let $\bm{x} = (x_1, \ldots, x_M)$ be the given utterance and $\bm{\theta}$ be the parameters of the model.
Our model is trained by updating $\bm{\theta}$ based on the loss for each task.

\paragraph{Generation}

The response to $\bm{x}$ is defined as $\bm{y} = (y_1, \ldots, y_N)$. The model infers an appropriate $\bm{y}$ from $\bm{x}$.
The generation loss $\mathcal{L}_\mathrm{gen}$ is calculated as the negative log-likelihood loss.
\begin{equation}
    \mathcal{L}_\mathrm{gen} = -\sum_{j=1}^N \log p (y_j | \bm{x}, y_1, \ldots, y_{j-1}; \bm{\theta})
    \label{eq:loss_gen}
\end{equation}

\paragraph{Classification}

If the correct label of $\bm{x}$ is $c$, the model infers $c$ from $\bm{x}$.
The negative log-likelihood loss is also used for the classification loss $\mathcal{L}_\mathrm{cls}$.
\begin{equation}
    \mathcal{L}_\mathrm{cls} = -\log p (c | \bm{x}; \bm{\theta})
    \label{eq:loss_cls}
\end{equation}

\subsection{Loss Weighting}

Although the proposed multi-task learning model learns the generation and classification tasks simultaneously, there is a possibility that the ratio of learning for the classification task is too large.
When solving a general classification task, the end of learning is often determined by the convergence of the loss in the validation data.
On the other hand, the target of our model is a generation task, and the number of epochs required for generation is larger than that of the classification task.

Therefore, we consider weighting the loss functions.
While the weight for response generation is fixed at 1, the weight for emotion recognition is varied between 0 and 1.
This makes the contribution of the classification task reduced in updating the parameters.

\begin{table}[t]
    \centering
    \begin{tabular}{lrrr}
        \hline
        \textbf{Dataset} & \textbf{Train} & \textbf{Validation} & \textbf{Test} \\
        \hline
        DailyDialog & 76,052 & 7,069 & 6,740 \\
        TEC & 16,841 & 2,105 & 2,105 \\
        SST-2 & 16,837 & 872 & 1,822 \\
        CrowdFlower & 15,670 & 1,958 & 1,958 \\
        \hline
    \end{tabular}
    \caption{The statistics of the datasets for our experiments, where TEC stands for Twitter Emotion Corpus.
             Because TEC and CrowdFlower have no split of train, validation, and test, we split them into three at 8:1:1.}
    \label{tab:data}
\end{table}

\begin{table*}[t]
    \centering
    \begin{tabular}{l|rrrr|rrrr}
        \hline
        \multirow{2}{*}{\textbf{Model}} & \multicolumn{4}{c|}{\textbf{Auto Eval}} & \multicolumn{4}{c}{\textbf{Manual Eval}} \\
        & \textbf{\textsc{Bleu}} & \textbf{\textit{dist}-1} & \textbf{\textit{dist}-2} & \textbf{Avg Len} & \textbf{\textit{Emo}} & \textbf{\textit{Flu}} & \textbf{\textit{Info}} & \textbf{\textit{Relv}} \\
        \hline
        R & 32.35 & 5.87 & 30.48 & 14.12 & 3.44 & 3.48 & 3.63 & 3.55 \\
        R+E6 & 32.29 & 5.93 & 30.48 & 14.12 & \textbf{3.59} & \textbf{3.82} & 3.62 & \textbf{3.96} \\
        R+E6+E2 & 32.39 & \textbf{6.00} & \textbf{30.77} & 14.11 & 3.58 & 3.75 & \textbf{3.74} & 3.70 \\
        R+E6+E12 & \textbf{32.55} & 5.89 & 30.57 & \textbf{14.14} & 3.52 & 3.48 & 3.55 & 3.58 \\
        R+E6+E2+E12 & 32.29 & 5.91 & 30.47 & 14.12 & \textbf{3.59} & 3.75 & 3.57 & 3.64 \\
        \hline
    \end{tabular}
    \caption{Evaluation results of our models by multi-task learning. R stands for response generation, and E$\bullet$ is emotion recognition with $\bullet$ labels. \textit{Emo}, \textit{flu}, \textit{info}, and \textit{relv} are the four aspects for the manual evaluation by crowdsourcing.}
    \label{tab:eval}
\end{table*}

\section{Experiments}

\subsection{Datasets}

We train a model with three tasks of emotion recognition in addition to response generation using multi-task learning.
Each emotion recognition task is a classification task with 6, 2, and 12 labels, and we call them emotion recognition, coarse-grained emotion recognition, and fine-grained emotion recognition, respectively.
The datasets for such emotion recognition were selected according to \citet{bostan-klinger-2018-analysis}.
The numbers of instances are summarized in Table~\ref{tab:data}.

\paragraph{Response Generation}

DailyDialog~\citep{li-etal-2017-dailydialog} is used for response generation.
The dataset is a multi-turn dialogue corpus, and we obtain pairs of an utterance and a response by extracting two turns at a time.
Each utterance in the corpus has an emotion label, but we do not use these labels in the experiment.
This is because almost all of the emotion labels are \texttt{other}, which is not suitable for our method.

\paragraph{Emotion Recognition}

For the core emotion recognition dataset, we use the Twitter Emotion Corpus~\citep{mohammad-2012-emotional}.
It was constructed based on Twitter hashtags and consists of six labels: \{\texttt{anger}, \texttt{disgust}, \texttt{fear}, \texttt{joy}, \texttt{sadness}, \texttt{surprise}\}.
Because there is no distinction between train, validation, and test in the dataset, 80\% of the total samples is assigned to train, and the remaining 10\% each is assigned to validation and test.

\paragraph{Coarse-Grained Emotion Recognition}

For coarse-grained emotion recognition, we use SST-2~\citep{socher-etal-2013-recursive}.
This is a dataset of movie comments labeled with \{\texttt{positive}, \texttt{negative}\}.
To maintain a balance with the number of instances for the other emotion recognition tasks, we reduce the number of instances for training to 25\%.

\paragraph{Fine-Grained Emotion Recognition}

For fine-grained emotion recognition, we use the emotionally-tagged corpus provided by CrowdFlower.\footnote{The original link is no longer available. An alternative is \url{https://data.world/crowdflower/sentiment-analysis-in-text}.}
We exclude the label \texttt{empty} and adopt this corpus for a classification task with 12 labels: \{\texttt{anger}, \texttt{boredom}, \texttt{enthusiasm}, \texttt{fun}, \texttt{happiness}, \texttt{hate}, \texttt{love}, \texttt{neutral}, \texttt{relief}, \texttt{sadness}, \texttt{surprise}, \texttt{worry}\}.
As with the Twitter Emotion Corpus, this corpus does not have a split of train, validation, and test, and thus the whole data is divided into 8:1:1.
Furthermore, for the same reason as in SST-2, only 50\% of the total data is used.

\subsection{Training}

The hyperparameters are set based on BART~\citep{lewis-etal-2020-bart} and the Fairseq example.\footnote{\url{https://github.com/pytorch/fairseq/blob/master/examples/bart/README.summarization.md}.}
The learning rate is set to 3e-5, and the parameters are optimized by Adam with weight decay.
For response generation, we apply label smoothing of 0.1 to the negative log-likelihood loss.
The number of input and output tokens is set to 64, and training is performed for 64 epochs.
We use beam search with 5 beams to select words and eliminate cases where there are more than three repeated $n$-grams.
Training and generation are performed on NVIDIA Tesla V100.

\begin{figure}[t]
    \centering
    \includegraphics[width=\linewidth]{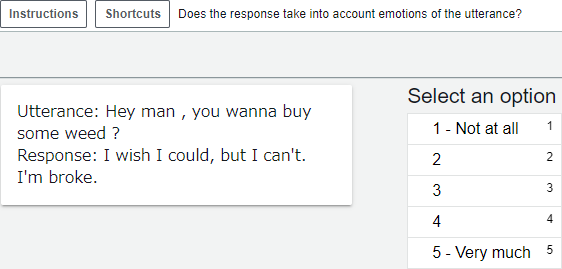}
    \caption{An example of the manual evaluation by crowdsourcing on Amazon Mechanical Turk.
             Workers are supposed to answer such questions by rating the given dialogue on a five-point scale.}
    \label{fig:mturk}
\end{figure}

\subsection{Evaluation Metrics}

We evaluate the trained models automatically and manually.

\paragraph{Automatic Evaluation}

First, we evaluate how much the output responses are related to the correct response using \textsc{Bleu}~\citep{papineni-etal-2002-bleu}.
Second, we evaluate whether the output responses are lexically diverse using \textit{distinct}~\citep{li-etal-2016-diversity}.
For \textit{distinct}, \textit{distinct}-1 and \textit{distinct}-2 are calculated, which focus on unigrams and bigrams, respectively.
We also compare the average number of words in output responses, which is based on the assumption that the longer a response is, the less common it is.
The large average number indicates that generated responses tend to be not dull.

\paragraph{Manual Evaluation}

Actually, the lack of correlation between automatic and manual evaluation~\citep{liu-etal-2016-evaluate} has been indicated especially in regards to generation tasks.
Thus, we perform manual evaluation by crowdsourcing, where Amazon Mechanical Turk is used as the platform.
We use four metrics mainly following \citet{rashkin-etal-2019-towards}: \textit{emotion}, \textit{fluency}, \textit{informativeness}, and \textit{relevance}.
Each of the questions asks whether the generated response takes into account the emotion of the utterance, whether the generated response is syntactically correct, whether a generated response provides some information for the utterance, and whether the content of the response is appropriately related to the utterance.
A total of 100 randomly selected responses for the test data are asked to rate the above four metrics on a five-point scale.
US residents are designated as workers, and seven workers are requested for each metric of each sample.
The final score is obtained as the average of the values obtained from the seven workers.
An example of the questions asked to the workers is shown in Figure~\ref{fig:mturk}.

\subsection{Results}

\begin{table*}[t]
    \centering
    \begin{tabular}{p{.22\linewidth}p{.22\linewidth}p{.22\linewidth}p{.22\linewidth}}
        \hline
        \textbf{Utterance} & \textbf{Gold Response} & \textbf{R} & \textbf{R+E6} \\
        \hline
        Yes, my parents, and soon my brother as well. & I didn't know you had a brother. & I see. In a few months, you'll be moving to a new place. & I hope I get to meet you soon. \\
        It falls on the fifth day of the fifth lunar month. & Could you tell me the origin of the two festivals? & Oh, then on that day? & Oh? Why is that day? \\
        How smart of you to book the tickets before hand! & Oh, do you really think so? & Sorry, I didn't think of it that way. & Yeah, yeah, I know. It was the last thing I wanted to do. \\
        I can't. I'm waiting for Paul, but he's late. & I'll say the movie's starting in the minute. & Would you like me to call him? & Where is he? \\
        thank you. I think I should also have a look at the Internet to see if anyone's got a sublet. & how much do are you looking to spend? & if you need a flat, you can always look online. Do you have any questions? & good idea. You can ask around at the office. If you need a flat, they'll probably have one. \\
        \hline
    \end{tabular}
    \caption{Examples generated by the baseline model (R) and the proposed model (R+E6).
             We chose examples where the \textit{emotion} score in manual evaluation is quite improved by adding E6 to R.}
    \label{tab:case}
\end{table*}

\begin{table}[t]
    \centering
    \begin{tabular}{lrr}
        \hline
        \textbf{Model} & \textbf{Acc [\%]} & \textbf{F1 [\%]} \\
        \hline
        R & 17.29 & 12.25 \\
        R+E6 & 64.32 & 56.53 \\
        R+E6+E2 & 63.23 & 54.88 \\
        R+E6+E12 & 66.27 & 57.86 \\
        R+E6+E2+E12 & 64.85 & 56.09 \\
        \hline
    \end{tabular}
    \caption{Emotion recognition (E6) performance of our models in Table \ref{tab:eval}.
             The values for R, trained only on response generation, are very low, while R+E6+E12 marks the best score among these models.}
    \label{tab:emo}
\end{table}

\paragraph{Multi-Task Learning}

The evaluation results are shown in Table~\ref{tab:eval}.
The response generation is denoted by R, and the emotion recognition for the Twitter Emotion Corpus, SST-2, and CrowdFlower datasets is denoted by E6, E2, and E12, respectively.
In terms of automatic evaluation, R+E6+E2 and R+E6+E12 maximized the \textit{distinct} and \textsc{Bleu}, respectively.
In the proposed multi-task learning model, therefore, emotion recognition of different granularity is effective in relevance and diversity.
For manual evaluation, all models that include emotion recognition outperformed the model with only response generation.
Moreover, R+E6 scores were particularly high for all four metrics.
The proposed multi-task learning model not only makes the generated responses more emotionally aware but can also improve the quality of other metrics, such as fluency and informativeness.

Several examples of responses generated by the obtained model are shown in Table~\ref{tab:case}.
We compare the given utterances and their responses of R and R+E6.
We can see that R+E6 generated more emotion-sensitive sentences, such as ``Yeah, yeah, I know'' and ``good idea.''

In addition, we show the results of emotion recognition in Table \ref{tab:emo}, which is especially on a six-label classification task.
We calculate accuracy and F1-score as metrics for evaluation.
The result shows that, on emotion recognition, increasing the number of tasks to train does not necessarily lead to improvement of the scores.
We can see that models with training of emotion recognition on fine-grained labels tend to outperform the other models.
However, the goal of our model is not improvement of classification but that of generation, so that those score variation is not essential in this work.

\paragraph{Loss Weighting}

The evaluation results for different loss weighting are shown in Table~\ref{tab:eval_weight}.
The weight for the loss of E$\bullet$ is denoted as $\lambda_{\mathrm{E}\bullet}$.
In automatic evaluation, we can see the improvement of the scores by weighting, especially in the model with E12.
On the other hand, the manual evaluation shows that weighting improves some scores, with the case (.5, .5, 0) producing the highest score.
Therefore, weighting each loss can improve the quality of generated responses, and in the condition of our experiment, it is most effective to reduce the weights of E6 and E2 by half.

\begin{table*}[t]
    \centering
    \begin{tabular}{l|rrrr|rrrr}
        \hline
        \multirow{2}{*}{$(\lambda_\mathrm{E6}, \lambda_\mathrm{E2}, \lambda_\mathrm{E12})$} & \multicolumn{4}{c|}{\textbf{Auto Eval}} & \multicolumn{4}{c}{\textbf{Manual Eval}} \\
        & \textbf{\textsc{Bleu}} & \textbf{\textit{dist}-1} & \textbf{\textit{dist}-2} & \textbf{Avg Len} & \textbf{\textit{Emo}} & \textbf{\textit{Flu}} & \textbf{\textit{Info}} & \textbf{\textit{Relv}} \\
        \hline
        (1, 0, 0) & 32.29 & 5.93 & 30.48 & 14.12 & 3.59 & 3.82 & 3.62 & \textbf{3.96} \\
        (.5, .5, 0) & 32.48 & 5.86 & 30.54 & \textbf{14.15} & \textbf{4.00} & \textbf{4.16} & \textbf{4.01} & \textbf{3.96} \\
        (.5, 0, .5) & \textbf{32.52} & 5.93 & 30.62 & 14.04 & 3.37 & 3.60 & 3.37 & 3.36 \\
        (.33, .33, .33) & 32.43 & \textbf{5.97} & \textbf{30.81} & 14.01 & 3.63 & 3.37 & 3.49 & 3.66 \\
        \hline
    \end{tabular}
    \caption{Evaluation results for differed loss. $\lambda_{\mathrm{E}\bullet}$ indicates the weight for the loss of E$\bullet$, and the metrics are the same as those of Table~\ref{tab:eval}. The weight for the response generation loss ($\lambda_\mathrm{R}$) is fixed at 1 throughout the experiments. Note that (1, 0, 0) is equivalent to R+E6 in Table~\ref{tab:eval}.}
    \label{tab:eval_weight}
\end{table*}

\section{Conclusion}

We worked on improving the quality of neural network-based response generation.
Focusing on the aspect of emotion, we proposed a multi-task learning response generation model that includes the tasks of generation and classification.
Through automatic and manual evaluations, we confirmed that the proposed model improved several metrics of performance.
Moreover, we further improved the quality of the model by weighting losses.
As a result, we found that such weighting improved several scores and the balance of parameter updates was also an important factor.

This paper focused on the emotion of the dialogue and generated responses that take into account the emotion of an utterance.
On the other hand, we did not focus on the emotion of a response, which is a subject for our future work.
We plan to work on estimating the emotions that a response should have and generating a response based on a specified emotion.
In the experiments of this paper, we omitted the context of a dialogue.
However, it is also necessary to consider past utterances and their effects on emotions for generating responses, which is also an issue to be addressed in the future.

\section*{Acknowledgements}

This work was supported by JSPS KAKENHI Grant Number JP18H03286.

\bibliography{anthology,custom}
\bibliographystyle{acl_natbib}

% \appendix

% \section{Example Appendix}
% \label{sec:appendix}

% This is an appendix.

\end{document}